\documentclass[11pt,a4paper]{article}

\usepackage[margin=1in]{geometry}
\usepackage[T1]{fontenc}
\usepackage[utf8]{inputenc}
\usepackage{lmodern}
\usepackage{microtype}
\usepackage{amsmath}
\usepackage{booktabs}
\usepackage{graphicx}
\usepackage{array}
\usepackage{caption}
\usepackage{fvextra}

\newcommand{\bmsubsection}[1]{\subsection*{#1}}
\usepackage{hyperref}
\usepackage[numbers,sort&compress]{natbib}
\usepackage{authblk}
\usepackage{needspace}

\hypersetup{colorlinks=true,linkcolor=blue,citecolor=blue,urlcolor=blue}

\title{When Should LLMs Abstain? Chain-of-Self-Questioning for Selective Risk Control}
\author{Ali \c{S}enol\\
\small Department of Computer Engineering, Tarsus University, Tarsus, T\"urkiye\\
\small ORCID: 0000-0003-0364-2837\\
\texttt{alisenol@tarsus.edu.tr}

}
\date{}

\begin{document}
\maketitle

\begin{abstract} Large language models can produce fluent answers when their factual support is weak. This paper introduces Chain-of-Self-Questioning (CoSQ), a prompt-only framework that makes answer commitment conditional on an explicit assessment of the information required to answer a question. We evaluate three CoSQ variants under seventeen conditions on the 817-item TruthfulQA multiple-choice validation set using eleven open-weight and hosted model families. In the final balanced-option protocol, Grounded-CoSQ at $\tau=0.90$ reduces the mean unconditional wrong-commitment rate from 13.1\% under chain-of-thought prompting to 8.9\%, a 32.1\% relative reduction, while increasing answered accuracy from 86.9\% to 89.7\% and answering 87.6\% of questions. Both improvements hold for all eleven models and at every evaluated threshold. Critical-CoSQ and Adaptive-CoSQ provide neighboring operating points with 88.6\% and 86.5\% coverage, respectively, while remaining more reliable than the baseline. A secondary Natural Questions Short-Answer evaluation provides convergent open-form evidence. These findings show that self-assessment can support explicit, tunable answer-or-abstain decisions when an unsupported commitment is more costly than referral or review.
\end{abstract}

\section{Introduction}

Large Language Models (LLMs) are increasingly used to answer factual questions, summarize evidence, and support decisions. Their fluency, however, can obscure an important limitation: a model may generate a plausible answer even when the information needed to support that answer is incomplete or wrong. This failure is commonly described as hallucination, namely the production of unsupported or factually incorrect content by a language-generation system \citep{ji2023survey,huang2025survey}. In high-consequence settings, including healthcare, law, finance, and public administration, an explicit abstention can be more useful than a confident but unsupported answer because it can trigger verification or human review. Recent work on domain-knowledge-enhanced LLM systems further illustrates that reliability becomes especially important when model outputs support fraud detection and concept-drift analysis rather than open-ended text generation \citep{senol2026domainknowledge}.

Most question-answering evaluations implicitly reward commitment. A model that guesses on every item may obtain a reasonable aggregate score even though it cannot distinguish answerable questions from questions for which it lacks reliable support. Selective prediction offers a different perspective: a system may reject some inputs in order to reduce risk on the inputs it answers \citep{chow1957optimum,elyaniv2010foundations,geifman2019selectivenet}. For LLMs, this perspective requires reporting answered accuracy together with coverage and the rate of wrong committed answers.

CoT prompting improves reasoning performance on many tasks, but it does not by itself create an answer-or-abstain decision. A model can reason coherently toward a false factual answer. Retrieval and post-hoc verification can improve reliability, but they require additional infrastructure or intervene after a candidate answer has already been produced \citep{lewis2020rag,manakul2023selfcheckgpt}. CoSQ addresses a narrower but complementary problem: whether the model should commit to an answer before it produces one. This distinction is especially important in high-stakes settings. A physician who refers a case to a specialist when the evidence is insufficient has not failed to provide care; the referral is a responsible decision that limits the risk of an incorrect diagnosis. CoSQ applies the same operational principle to language-model answers: when the evidence for a commitment is insufficient, abstention can be preferable to a fluent but unsupported response.

This paper presents CoSQ as a model-agnostic three-stage prompting
framework. First, the model identifies the information units needed to
answer the question. Second, it evaluates the support for those units.
Third, it either answers using the accepted information or abstains.
Grounded-CoSQ is the framework's primary risk-oriented
instantiation in this study, whereas Critical-CoSQ and Adaptive-CoSQ
provide neighboring selective policies. At the prespecified
$\tau=0.90$ operating point, Grounded-CoSQ yields the highest mean
answered accuracy, retains more coverage than Adaptive-CoSQ, and has a
wrong-commitment rate within 0.03 percentage points of the panel minimum.
It is therefore treated as the principal balanced configuration rather
than as a universally superior variant. Its final answer is conditioned on the information units that pass the gate, so the method makes the connection between self-assessment and answer generation explicit. The confidence threshold is not treated as a calibrated probability; it is an empirical operating parameter that determines a point on a risk--coverage frontier.

The paper makes four contributions:
\begin{enumerate}
  \item It introduces CoSQ, a prompt-only selective answering framework that separates information assessment from answer commitment.
  \item It evaluates the framework across eleven model families on a deterministic multiple-choice TruthfulQA protocol with balanced answer-option positions, avoiding both open-form parsing ambiguity and a fixed-label shortcut.
  \item It reports the complete threshold sweep and compares grounded, critical-item, and adaptive variants rather than presenting a single post-hoc threshold.
  \item It analyzes the method using answered accuracy, coverage, hallucination rate, paired model-level tests, confidence intervals, and model-level visualizations.
\end{enumerate}

\subsection{Hypotheses and research questions}

The evaluation is organized around two directional hypotheses and two
research questions:

\begin{description}
  \item[H1 (risk reduction):]
  At the primary operating point ($\tau=0.90$), Grounded-CoSQ produces
  a lower unconditional wrong-commitment rate than forced-choice
  chain-of-thought prompting.

  \item[H2 (conditional reliability):]
  Among committed answers, Grounded-CoSQ at $\tau=0.90$ produces higher
  answered accuracy than forced-choice chain-of-thought prompting.

  \item[RQ1 (operating-point control):]
  How does the confidence threshold affect answered accuracy, coverage,
  and the unconditional wrong-commitment rate across the Grounded,
  Critical, and Adaptive CoSQ variants?

  \item[RQ2 (generalization):]
  How consistently do the observed effects generalize across model
  families and factual question-answering settings?
\end{description}

H1 and H2 are evaluated on TruthfulQA-MC. RQ1 is examined through the
complete threshold sweep, whereas RQ2 is investigated using model-level
results and the secondary Natural Questions Short-Answer evaluation.

The remainder of the paper is organized as follows. Section~2 reviews work on hallucination, uncertainty, grounding, and selective prediction. Section~3 defines the CoSQ variants, and Section~4 describes the experimental design. Section~5 presents the results, followed by discussion, limitations, and conclusions in Sections~6--8. The appendices provide the prompt templates and supplementary analyses.

\section{Related Work}

\subsection{Hallucination and truthfulness}
Hallucination in natural language generation includes unsupported or factually incorrect content and is influenced by data conflicts, decoding, knowledge limitations, and failures of grounding \citep{ji2023survey,huang2025survey}. TruthfulQA was designed to test whether models reproduce common misconceptions rather than provide truthful answers \citep{lin2022truthfulqa}. Its multiple-choice formulation is particularly useful for this study because correctness can be scored without a fragile open-form semantic parser.

Evaluation design also changes the incentives created by a benchmark. Recent work argues that optimizing ordinary accuracy can encourage models to guess rather than abstain \citep{kalai2026accuracy}. This paper therefore treats abstention as an observable decision and does not interpret a lower coverage operating point as a failure without also reporting its risk. More generally, recent evaluation work argues that LLM quality should be assessed across multiple behavioral dimensions rather than by final-answer accuracy alone, including consistency, robustness, and reliability under changing conditions \citep{senol2026rqeval}.

\subsection{Reasoning, uncertainty, and self-knowledge}
CoT prompting elicits intermediate reasoning \citep{wei2022cot}, while self-consistency aggregates multiple reasoning paths \citep{wang2023selfconsistency}. These approaches primarily improve answer generation. They do not necessarily determine whether the model should answer at all. Research on calibration, verbalized uncertainty, and latent knowledge suggests that models sometimes expose useful signals about their own uncertainty, although these signals are often imperfect \citep{guo2017calibration,jiang2021know,kadavath2022mostly,lin2022uncertainty,tian2023calibration,xiong2024uncertainty,burns2023latent}. CoSQ operationalizes this signal as a selective decision rather than using confidence only as explanatory text. Recent behavioral research has likewise proposed measuring epistemic honesty through observable answer, abstention, and confidence patterns, while explicitly separating such behavioral evidence from claims about human-like introspective access \citep{senol2026epistemichonesty}.

This operationalization should not be confused with a claim that the model possesses human-like introspective access to its own knowledge. In this paper, ``self-questioning'' denotes a prompt-level procedure that elicits a confidence signal and converts it into an answer-or-abstain decision. The scientific claim is therefore behavioral and selective: the elicited signal is useful when it changes the risk--coverage profile, even if it is not a calibrated probability or a direct measurement of metacognitive awareness.

Selective generation and abstention methods have explored self-evaluation and trained rejection behavior \citep{ren2023selfevaluation,amayuelas2024knowledge,zhang2024rtuning,feng2024abstain}. These methods motivate the present work, but commonly require additional training, multiple models, or task-specific resources. CoSQ is deliberately prompt-only and is intended for black-box or hosted models.

\subsection{Grounding, verification, and selective prediction}
Retrieval-augmented generation grounds answers in external passages but introduces retrieval and evidence-selection failure modes \citep{lewis2020rag}. Black-box detectors such as SelfCheckGPT compare sampled answers after generation \citep{manakul2023selfcheckgpt}, while semantic uncertainty methods estimate uncertainty from meaning variation \citep{kuhn2023semantic}. CoSQ is complementary: it performs an explicit pre-commitment assessment and can be used before retrieval, after retrieval, or alongside post-hoc verification.

The risk--coverage formulation has a long history in reject-option classification \citep{chow1957optimum,chow1970optimum,elyaniv2010foundations,geifman2019selectivenet}. We adapt that formulation to LLM prompting. The core distinction is between answered accuracy, which is conditional on commitment, and coverage, which measures how often the system commits. Neither is sufficient by itself.

The present paper focuses on selective factual answering and does not assume that methods developed for unrelated clustering or streaming tasks transfer directly to language-model uncertainty.

\section{Chain-of-Self-Questioning}

\begin{figure*}[t]
\centering
\includegraphics[width=0.98\linewidth]{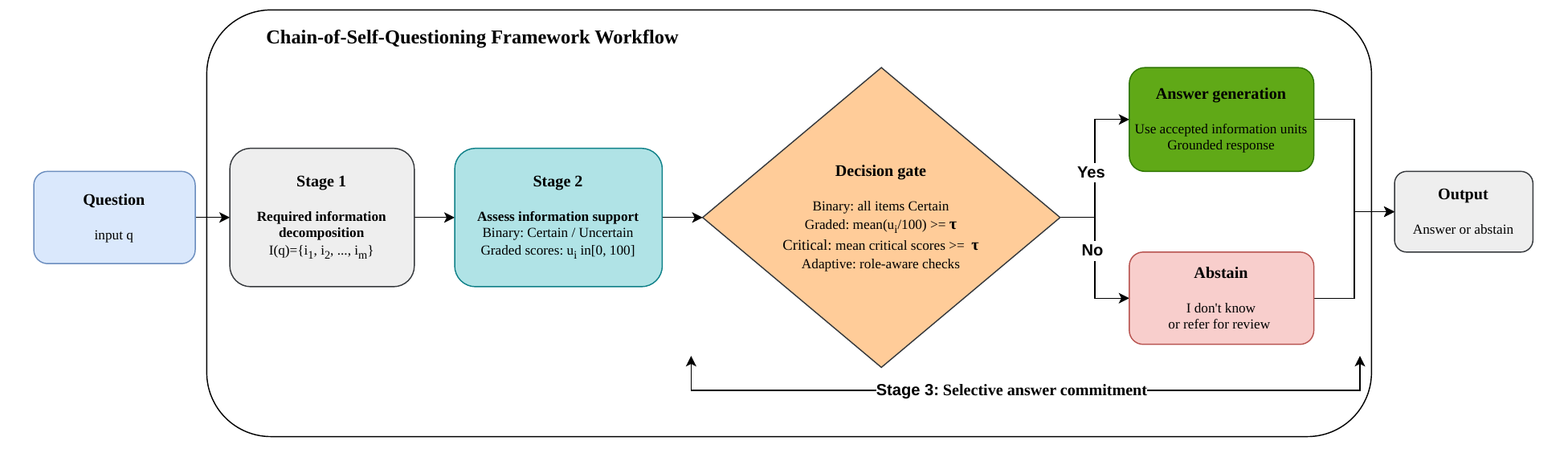}
\caption{Chain-of-Self-Questioning. The model identifies required information, evaluates its support, applies a confidence gate, and either abstains or generates an answer using accepted information units.}
\label{fig:workflow}
\end{figure*}

Figure~\ref{fig:workflow} summarizes the framework. Given a question $q$, the model first constructs information units $I(q)=\{i_1,\ldots,i_m\}$. It then assigns a confidence score $s_j\in[0,1]$ to each unit. A decision rule determines whether the model is allowed to commit. In the grounded variant, only accepted units are passed to the answer stage, which reduces the chance that the final answer is generated from an explicitly rejected premise.

\subsection{Grounded-CoSQ}

The grounded score is the mean of the item-level scores:
\begin{equation}
    \bar{s}(q)=\frac{1}{m}\sum_{j=1}^{m}s_j.
\end{equation}
The system commits when $\bar{s}(q)\geq\tau$ and abstains otherwise. If the gate is passed, the final prompt receives the accepted information units and asks for a concise answer. Thresholds $\tau\in\{0.50,0.60,0.70,0.80,0.90\}$ are evaluated before selecting the primary conservative operating point. We call the selected condition Grounded-CoSQ ($\tau=0.90$), not ``mean90'', in the manuscript.

\subsection{Critical-CoSQ}

Critical-CoSQ extends the information-unit stage by asking the model to label each unit as either critical or supporting. Only critical units participate in the gate. This reduces the mechanical effect of the number of supporting details and tests whether selective answering improves when the decision rule focuses on information without which a correct answer is impossible. Critical-CoSQ ($\tau=0.90$) is reported as the principal coverage-oriented comparison.

\subsection{Adaptive-CoSQ}

Adaptive-CoSQ retains role-aware information units and applies three simultaneous checks. Let $\bar{s}$ be the mean confidence across all claims, $\bar{s}_{c}$ the mean confidence across critical claims, and $s_{c,\min}$ the minimum critical-claim confidence. For each value in the prespecified sweep $\tau\in\{0.50,0.60,0.70,0.80,0.90\}$, the question is accepted only when $\bar{s}\geq\tau$, $\bar{s}_{c}\geq0.65$, and $s_{c,\min}\geq0.40$. Thus, $\tau$ controls the overall-confidence gate, whereas the critical-mean and minimum-critical checks remain fixed. After acceptance, claims with confidence at least 0.40 are passed to the answer stage. The answer is generated in a confident mode when $\bar{s}\geq0.75$ and in a cautious mode otherwise, then checked for contradiction; a detected contradiction produces an abstention. All thresholds and checks were fixed before aggregation.

\subsection{Baselines}

Direct prompting asks for the answer without an explicit reasoning or abstention stage. CoT asks the model to reason before answering. These baselines measure the cost of adding a selective gate. The comparison is intentionally prompt-level: no method receives task-specific fine-tuning or access to model logits.

\section{Experimental Design}

\subsection{Dataset and scoring protocol}

The primary benchmark is the 817-question validation split of TruthfulQA in its multiple-choice configuration \citep{lin2022truthfulqa}. Each question contains a fixed set of answer options and one gold answer. To remove the fixed-position structure of the source cache, option order is deterministically balanced with seed 1002 before prompting. Within each option-count stratum, the gold label frequencies differ by at most one whenever the number of questions permits exact balancing. A response is correct when its selected label matches the balanced gold label. Explicit CoSQ abstention is a separate outcome and does not enter the denominator of answered accuracy. Invalid CoSQ outputs are retained as unparseable. Direct and CoT are forced-choice baselines: they have coverage 1 by design, and a malformed response is conservatively scored as wrong rather than reclassified as an abstention.

The study also includes a 300-question exploratory Natural Questions Short-Answer (NQ-Short) evaluation based on the Natural Questions benchmark \citep{kwiatkowski2019natural}. NQ-Short is reported in the main Results section as a secondary generalization test. The confirmatory TruthfulQA-MC analysis remains primary because it provides a common deterministic scoring rule across all model conditions.

\subsection{Model panel}

The full panel contains eleven contemporary instruction-tuned or hosted model families, summarized in Table~\ref{tab:models}. Endpoint labels are preserved exactly as recorded in the final run manifests. Hosted labels do not necessarily expose immutable weight revisions; this is reported as a reproducibility limitation rather than treated as evidence of a particular model version.

\begin{table*}[t]
\centering
\caption{Full TruthfulQA-MC model panel. The recorded endpoint labels are preserved from the experiment manifests; NQ denotes the secondary Natural Questions analysis.}
\label{tab:models}
\small
\begin{tabular}{p{0.23\linewidth}p{0.19\linewidth}p{0.14\linewidth}p{0.18\linewidth}p{0.13\linewidth}}
\toprule
Recorded endpoint label & Family/class & Parameter class & Access class & Dataset use \\
\midrule
llama3-8b & Llama 3 & 8B & open-weight & TQA, NQ \\
llama3-70b & Llama 3 & 70B & open-weight & TQA \\
llama4\_scout-17b & Llama 4 Scout & 17B & open-weight & TQA, NQ \\
gemma3\_12b\_it & Gemma 3 instruction-tuned & 12B & open-weight & TQA \\
gemma4\_31b\_it & Gemma 4 instruction-tuned & 31B & open-weight & TQA, NQ \\
mistral-7b & Mistral & 7B & open-weight & TQA \\
gpt-oss-20b & GPT-OSS & 20B & open-weight & TQA, NQ \\
gpt-oss-120b & GPT-OSS & 120B & open-weight & TQA \\
gpt5\_5 & GPT-5.5 & not disclosed & hosted & TQA, NQ \\
claude5\_sonnet & Claude 5 Sonnet & not disclosed & hosted & TQA \\
deepseek-flash & DeepSeek Flash & not disclosed & hosted & TQA \\
\bottomrule
\end{tabular}
\end{table*}

\subsection{Conditions and reproducibility}

Each model is evaluated under seventeen conditions: Direct and CoT, together with Grounded-CoSQ, Critical-CoSQ, and Adaptive-CoSQ at $\tau=0.50,0.60,0.70,0.80,0.90$. The same 817 questions, balanced option order, generation settings, and prompt versions are used across the eleven models. The final manifests share one question-ID hash, one balanced-option hash, and one prompt digest. Model responses and intermediate decisions are cached, and aggregate results are generated from the cached records. This prevents accidental re-querying and permits independent re-analysis of scoring rules.

The framework, configuration examples, and analysis utilities are publicly available in the CoSQ repository and PyPI package. The experiment-specific prompt templates are supplied with the supplementary reproducibility package. Provider credentials, private endpoints, and local caches are excluded.

\subsection{Metrics}

For $N$ questions, let $C$ denote correct committed answers, $W$ wrong committed answers, and $A$ abstentions. We report:
\begin{align}
AA &= \frac{C}{C+W}, & Coverage &= \frac{C+W}{N},\\
HR &= \frac{W}{N}, & AR &= \frac{A}{N},\\
\mathrm{Accuracy}_{\mathrm{overall}} &= \frac{C}{N}.
\end{align}
Answered Accuracy (AA) is the primary conditional correctness measure for committed answers. It must always be read together with coverage. Here Hallucination Rate (HR) denotes the unconditional wrong-commitment rate: in the multiple-choice protocol it is the proportion of all questions for which the model commits to an incorrect option. It should not be interpreted as a complete linguistic annotation of every type of hallucination. We also report the conditional answered error rate, $R_{answered}=W/(C+W)=1-AA$, which is the conventional selective risk among committed answers. For parseable outputs, $HR=Coverage\times(1-AA)$.

\subsection{Statistical analysis}

The primary comparisons are paired across the eleven models because every model is evaluated under the same question set and conditions. For H1, we compute the model-level reduction in HR from CoT to each CoSQ variant; for H2, we compute the corresponding increase in AA. The directional hypotheses are evaluated with exact one-sided Wilcoxon signed-rank tests. We also report percentile bootstrap intervals based on 10,000 resamples of the paired model-level differences and paired-sample Cohen's $d_z$ as a standardized descriptive effect size. Because the model count is small and model endpoints are not independent training examples, inferential results are interpreted alongside per-model values and complete threshold curves. The threshold sweep is reported in full to avoid selecting an operating point from a hidden test result.

\subsection{Inference cost}
CoSQ adds model calls because information extraction and confidence assessment precede answer generation. The relative stage count in the implementation is summarized in Table~\ref{tab:cost}. This overhead is a deliberate exchange: the system spends additional inference budget to reduce the probability of a wrong committed answer. Exact provider costs are excluded because they depend on the serving endpoint and are not part of the scientific comparison.

\begin{table*}[t]
\centering
\caption{Relative inference stages per question. A stage denotes one prompt--completion interaction in the CoSQ pipeline.}
\label{tab:cost}
\small
\begin{tabular}{lrl}
\toprule
Condition & Relative stages & Operational role \\
\midrule
Direct & 1 & Answer generation \\
CoT & 1 & Reasoning and answer generation \\
Grounded-CoSQ & 3 & Information needs, confidence, gated answer \\
Critical-CoSQ & 3 & Role classification, critical confidence, gated answer \\
Adaptive-CoSQ & 3--4 & Role-aware confidence and adaptive answer \\
\bottomrule
\end{tabular}
\end{table*}

\section{Results}

\subsection{Main model-panel results}

Table~\ref{tab:main} reports the model-macro mean and standard deviation across all eleven models. As required by the forced-choice protocol, Direct and CoT both have 100\% coverage. CoT attains 86.9\% AA and a 13.1\% unconditional wrong-commitment rate; it does not improve on the Direct baseline in this setting. Grounded-CoSQ at $\tau=0.90$ reduces HR to 8.9\% and increases AA to 89.7\% while answering 87.6\% of questions. At the same threshold, Critical-CoSQ answers 88.6\% of questions with 9.2\% HR, whereas Adaptive-CoSQ attains 89.6\% AA, 86.5\% coverage, and 8.9\% HR.

\begin{table*}[t]
\centering
\caption{Final balanced-option TruthfulQA-MC results across eleven models and 817 questions. Values are model-macro mean $\pm$ standard deviation. HR is the unconditional wrong-commitment rate $W/N$.}
\label{tab:main}
\small
\begin{tabular}{lrrrrr}
\toprule
Condition & AA & Coverage & HR & Abstention & Unparseable \\
\midrule
Direct & $0.872\pm0.033$ & $1.000\pm0.000$ & $0.128\pm0.033$ & $0.000\pm0.000$ & $0.000\pm0.000$ \\
CoT & $0.869\pm0.038$ & $1.000\pm0.000$ & $0.131\pm0.038$ & $0.000\pm0.000$ & $0.000\pm0.000$ \\
Grounded-CoSQ $\tau=0.50$ & $0.885\pm0.036$ & $0.945\pm0.020$ & $0.109\pm0.037$ & $0.054\pm0.020$ & $0.001\pm0.001$ \\
Grounded-CoSQ $\tau=0.60$ & $0.886\pm0.037$ & $0.944\pm0.021$ & $0.108\pm0.038$ & $0.056\pm0.022$ & $0.001\pm0.001$ \\
Grounded-CoSQ $\tau=0.70$ & $0.886\pm0.037$ & $0.943\pm0.021$ & $0.108\pm0.038$ & $0.056\pm0.022$ & $0.001\pm0.001$ \\
Grounded-CoSQ $\tau=0.80$ & $0.888\pm0.035$ & $0.937\pm0.019$ & $0.105\pm0.035$ & $0.062\pm0.020$ & $0.001\pm0.001$ \\
\textbf{Grounded-CoSQ $\tau=0.90$} & $\mathbf{0.897\pm0.021}$ & $0.876\pm0.058$ & $0.089\pm0.013$ & $0.123\pm0.058$ & $0.001\pm0.001$ \\
Critical-CoSQ $\tau=0.90$ & $0.896\pm0.021$ & $0.886\pm0.058$ & $0.092\pm0.016$ & $0.113\pm0.058$ & $0.001\pm0.001$ \\
Adaptive-CoSQ $\tau=0.90$ & $0.896\pm0.026$ & $0.865\pm0.047$ & $0.089\pm0.017$ & $0.134\pm0.047$ & $0.000\pm0.001$ \\
\bottomrule
\end{tabular}
\end{table*}

A raw-completion audit confirms that 8,936 of 8,987 outputs (99.43\%) in each forced baseline contain exactly one option label. The remaining 51 Direct and 51 CoT outputs violate the requested format and are scored as wrong. No malformed forced-choice output is treated as an abstention or credited as correct; consequently, their 100\% coverage reflects the evaluation contract rather than perfect instruction compliance.

\subsection{Abstention quality and lost value}

Risk reduction is not sufficient to establish practical usefulness because a selective system may abstain on questions that its baseline would have answered correctly. We therefore paired each $\tau=0.90$ abstention with the corresponding CoT outcome. A ``correct abstention'' is an abstention on a question where CoT was incorrect; ``lost value'' is an abstention on a question where CoT was correct. The percentages in Table~\ref{tab:abstention} are averaged across the eleven models and describe the composition of abstentions, not a second accuracy definition.

\begin{table*}[t]
\centering
\caption{Abstention composition at $\tau=0.90$. Values are means across eleven models; shares are conditional on the questions on which the corresponding variant abstained.}
\label{tab:abstention}
\small
\begin{tabular}{lrrr}
\toprule
Variant & Mean abstentions & Correct abstention share & Lost value share \\
\midrule
Grounded-CoSQ & 100.5 & 0.313 & 0.687 \\
Critical-CoSQ & 92.7 & 0.360 & 0.640 \\
Adaptive-CoSQ & 109.5 & 0.344 & 0.656 \\
\bottomrule
\end{tabular}
\end{table*}

\begin{figure*}[t]
\centering
\includegraphics[width=0.98\linewidth]{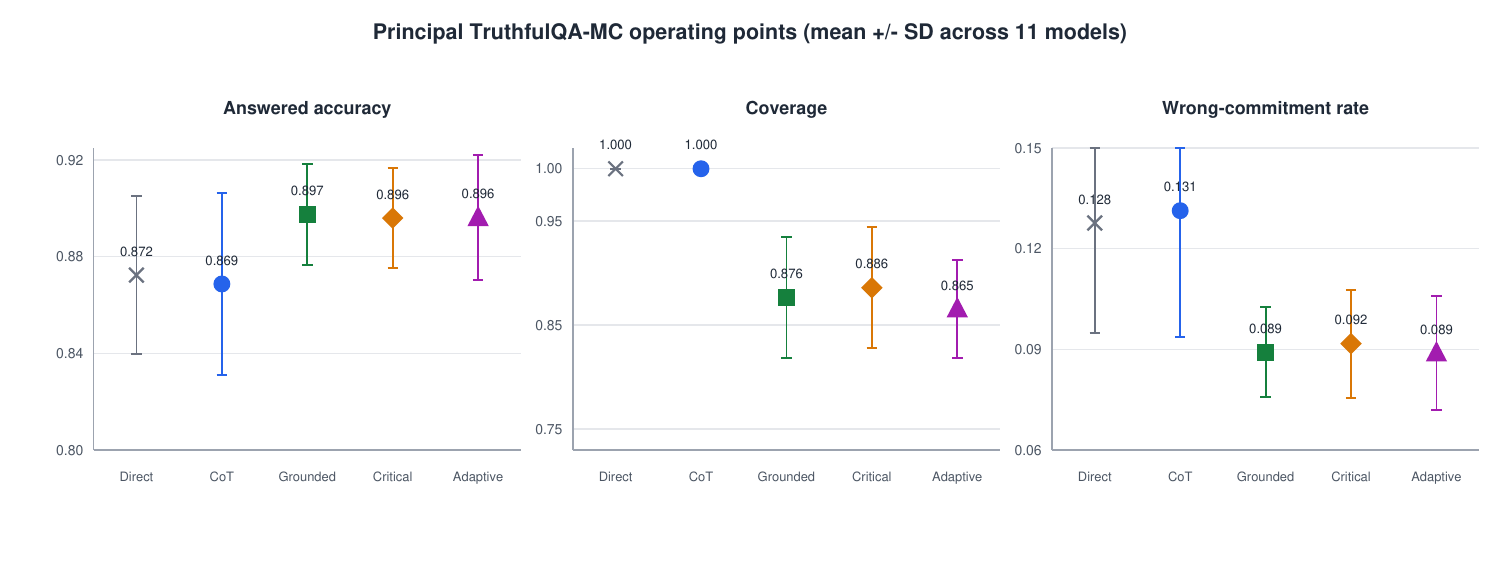}
\caption{Principal TruthfulQA-MC outcomes across the eleven-model panel. Points show model-macro means and error bars show one standard deviation. Colors and marker shapes identify Direct, CoT, and the three CoSQ variants consistently across figures.}
\label{fig:mainmetrics}
\end{figure*}

The abstention profile quantifies what each operating policy refers for review. Adaptive-CoSQ is the most conservative by response volume, whereas Critical-CoSQ retains the broadest coverage. Grounded-CoSQ occupies a narrow middle position: it refers 100.5 questions per model on average, and 31.3\% of those referrals coincide with a CoT error. The remaining referrals withhold answers that CoT happens to answer correctly. They are not scoring errors, but they quantify the reduction in automated service breadth associated with selective commitment. These values are operating characteristics, not evidence that abstention itself is a failure. Figure~\ref{fig:mainmetrics} compares the principal operating points across all three reported metrics.

The principal result is a consistent reduction in wrong commitments accompanied by higher conditional accuracy. Compared with CoT, Grounded-CoSQ $\tau=0.90$ reduces mean HR by 4.22 percentage points, a 32.1\% relative reduction, and increases AA by 2.87 points while directing 12.3\% of questions to explicit abstention. Critical-CoSQ reduces HR by 3.96 points and increases AA by 2.73 points while abstaining on 11.3\% of questions. Adaptive-CoSQ reduces HR by 4.24 points and increases AA by 2.75 points while abstaining on 13.4\%. These results define distinct but closely spaced operating profiles rather than a single ordering. Grounded-CoSQ has the highest mean AA and more coverage than Adaptive-CoSQ, whose mean HR is lower by only 0.02 percentage points; Critical-CoSQ preserves the most coverage among the $\tau=0.90$ variants.

\subsection{Threshold sweep}

\begin{figure*}[t]
\centering
\includegraphics[width=0.98\linewidth]{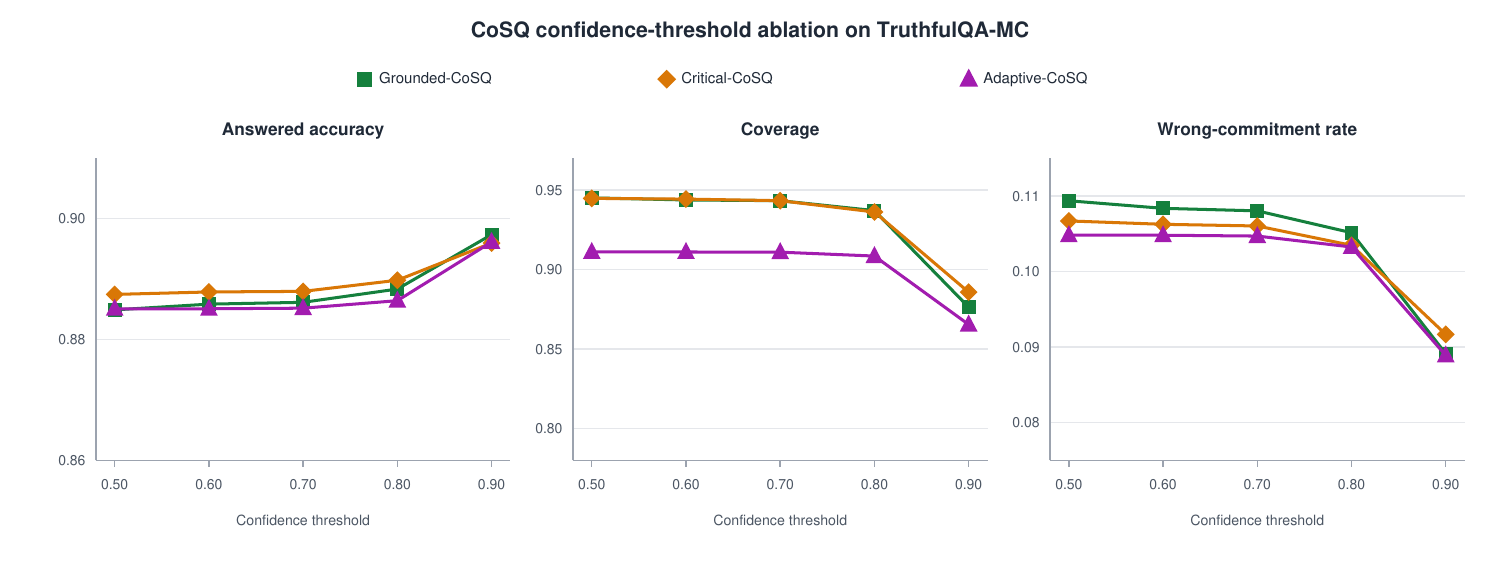}
\caption{Confidence-threshold ablation for the three CoSQ variants on TruthfulQA-MC. Panels report answered accuracy, coverage, and unconditional wrong-commitment rate. Colors and marker shapes distinguish the variants; all horizontal axes show $\tau$ from 0.50 to 0.90.}
\label{fig:threshold}
\end{figure*}

Figure~\ref{fig:threshold} shows the complete sweep for all three CoSQ variants. Every reported threshold produces a lower mean HR and higher mean AA than CoT. Moreover, both directions are favorable for every model under all fifteen selective configurations. Increasing $\tau$ generally lowers both coverage and HR, while the largest AA gains occur at $\tau=0.90$. The nearly identical values at some lower thresholds reflect concentration of the elicited confidence scores rather than omitted conditions. The threshold is therefore an explicit deployment control whose appropriate value depends on the relative consequences of a wrong commitment and a referral.

For completeness, Table~\ref{tab:variantssweep} reports the aggregate threshold values for the Critical and Adaptive variants. These values are not used to select the primary operating point; they document the full prespecified condition set.

\begin{table*}[t]
\centering
\caption{Aggregate threshold sweep for Critical-CoSQ and Adaptive-CoSQ on TruthfulQA-MC. Values are means across the eleven models.}
\label{tab:variantssweep}
\small
\begin{tabular}{lrrr}
\toprule
Condition & AA & Coverage & HR \\
\midrule
Critical-CoSQ $\tau=0.50$ & 0.887 & 0.945 & 0.107 \\
Critical-CoSQ $\tau=0.60$ & 0.888 & 0.944 & 0.106 \\
Critical-CoSQ $\tau=0.70$ & 0.888 & 0.943 & 0.106 \\
Critical-CoSQ $\tau=0.80$ & 0.890 & 0.936 & 0.103 \\
Critical-CoSQ $\tau=0.90$ & 0.896 & 0.886 & 0.092 \\
Adaptive-CoSQ $\tau=0.50$ & 0.885 & 0.911 & 0.105 \\
Adaptive-CoSQ $\tau=0.60$ & 0.885 & 0.911 & 0.105 \\
Adaptive-CoSQ $\tau=0.70$ & 0.885 & 0.911 & 0.105 \\
Adaptive-CoSQ $\tau=0.80$ & 0.886 & 0.909 & 0.103 \\
Adaptive-CoSQ $\tau=0.90$ & 0.896 & 0.865 & 0.089 \\
\bottomrule
\end{tabular}
\end{table*}

\subsection{Cross-dataset generalization: NQ-Short}

To test whether the observed behavior is specific to adversarial TruthfulQA questions, we additionally evaluated a 300-question short-answer subset of Natural Questions using five representative models. Unlike TruthfulQA-MC, NQ-Short uses open-form answers and therefore has a different semantic scoring protocol. We report it as a secondary generalization analysis rather than combine it numerically with the primary MC results.

\begin{figure*}[t]
\centering
\includegraphics[width=\linewidth]{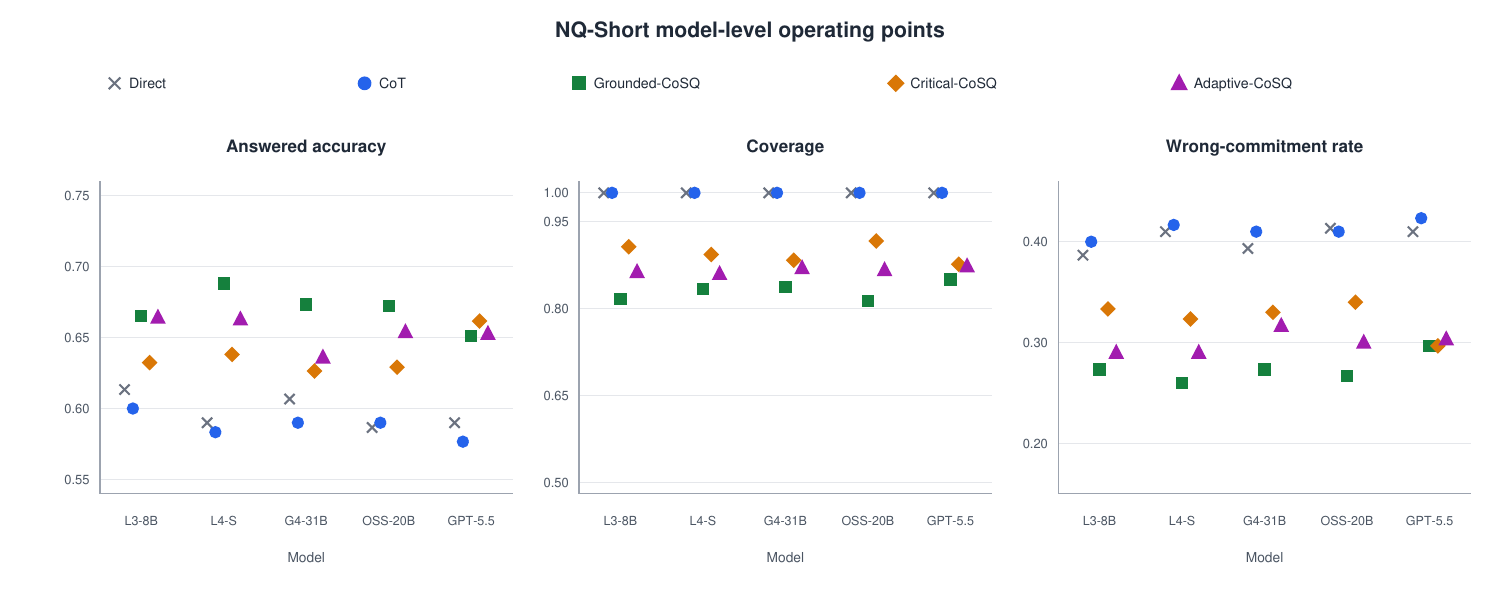}
\caption{Model-level NQ-Short results for the five-model secondary panel. Each panel reports one metric; colors identify the five evaluation conditions. The figure is descriptive and is not pooled with the primary TruthfulQA-MC analysis.}
\label{fig:nqmodel}
\end{figure*}

Table~\ref{tab:nq} reports the prespecified current five-model panel at the conservative $\tau=0.90$ operating point. Grounded-CoSQ reduces mean HR from 0.412 under CoT to 0.274 while retaining 0.830 coverage and increasing AA from 0.588 to 0.670. Critical-CoSQ preserves the broadest coverage (0.895) at a higher HR of 0.325, whereas Adaptive-CoSQ provides an intermediate operating point (0.867 coverage, 0.300 HR). Grounded-CoSQ improves both AA and HR relative to CoT for all five models. Figure~\ref{fig:nqmodel} visualizes these model-level outcomes, and Appendix~\ref{app:nqmodel} provides their exact values. These values should not be compared numerically with the primary MC estimates because NQ uses open-form generation and semantic answer matching, but they provide convergent evidence that selective risk control transfers beyond TruthfulQA-MC.

\begin{table*}[t]
\centering
\caption{Secondary Natural Questions Short-Answer results across five models and 300 questions. Values are mean $\pm$ standard deviation across models. HR is the unconditional wrong-commitment rate $W/N$.}
\label{tab:nq}
\small
\begin{tabular}{lrrrrr}
\toprule
Condition & AA & Coverage & HR & Abstention & Unparseable \\
\midrule
Direct & $0.597\pm0.012$ & $1.000\pm0.000$ & $0.403\pm0.012$ & $0.000\pm0.000$ & $0.000\pm0.000$ \\
CoT & $0.588\pm0.009$ & $1.000\pm0.000$ & $0.412\pm0.009$ & $0.000\pm0.000$ & $0.000\pm0.000$ \\
Grounded-CoSQ $\tau=0.90$ & $0.670\pm0.013$ & $0.830\pm0.015$ & $0.274\pm0.014$ & $0.170\pm0.015$ & $0.000\pm0.000$ \\
Critical-CoSQ $\tau=0.90$ & $0.638\pm0.014$ & $0.895\pm0.016$ & $0.325\pm0.017$ & $0.105\pm0.016$ & $0.000\pm0.000$ \\
Adaptive-CoSQ $\tau=0.90$ & $0.654\pm0.011$ & $0.867\pm0.005$ & $0.300\pm0.011$ & $0.133\pm0.005$ & $0.000\pm0.000$ \\
\bottomrule
\end{tabular}
\end{table*}

With respect to RQ2, the cross-dataset results provide cautious evidence
that the selective behavior generalizes beyond the primary benchmark. Grounded-CoSQ has the lowest HR and highest AA among the selective NQ conditions, while Critical-CoSQ preserves the most coverage. Ordinary factual questions in NQ-Short therefore yield a useful operating-point trade-off, but not a universal threshold recommendation. Because NQ-Short uses open-form semantic scoring and a smaller model panel, this finding should be treated as convergent evidence rather than a second confirmatory estimate.

\subsection{Model-level consistency}

\begin{figure*}[t]
\centering
\includegraphics[width=0.98\linewidth]{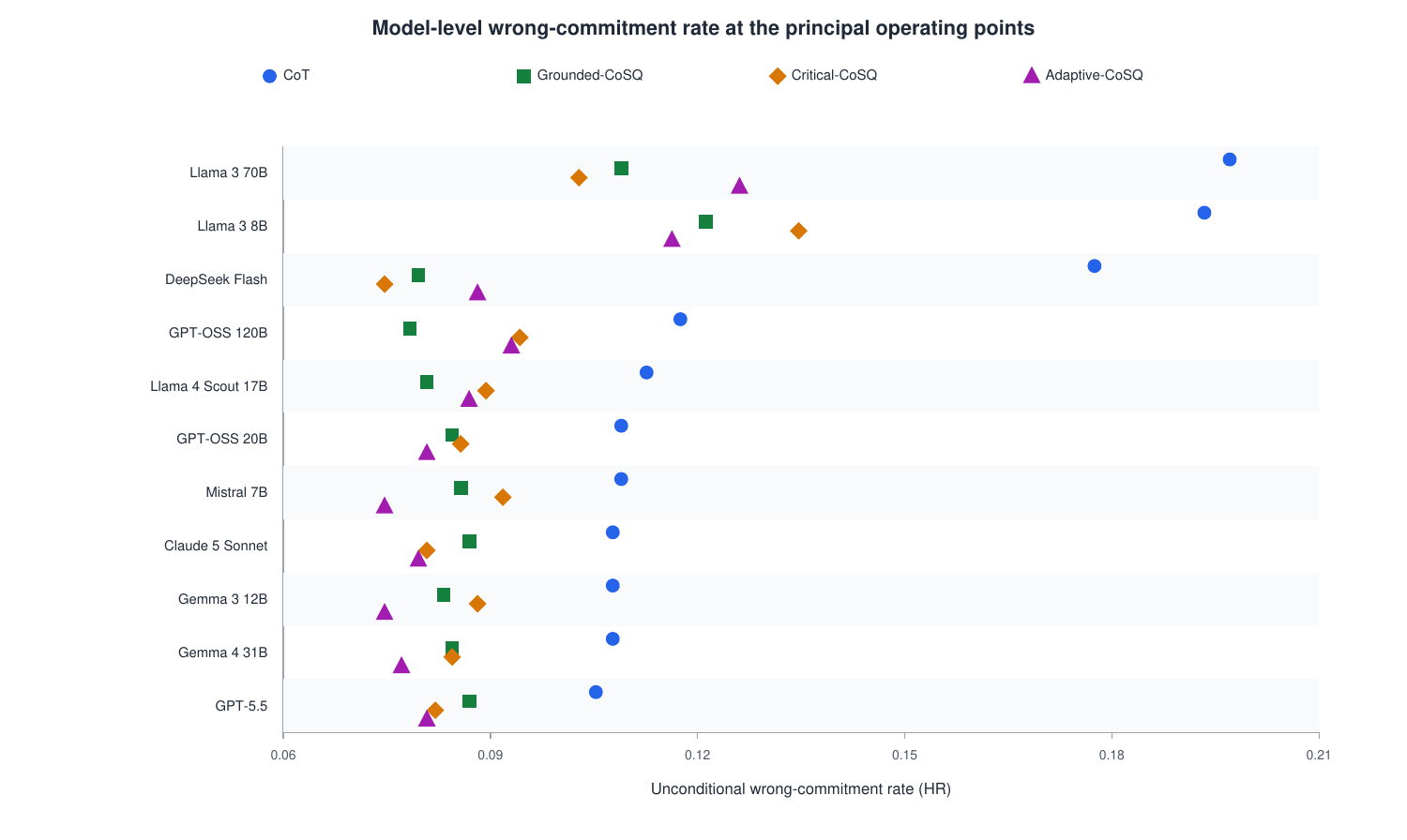}
\caption{Model-level unconditional wrong-commitment rate for CoT and the three $\tau=0.90$ CoSQ variants. Lower values are better. Colors and marker shapes have the same meaning as in Figures~\ref{fig:mainmetrics}--\ref{fig:threshold}.}
\label{fig:modelhr}
\end{figure*}

\begin{figure*}[t]
\centering
\includegraphics[width=0.98\linewidth]{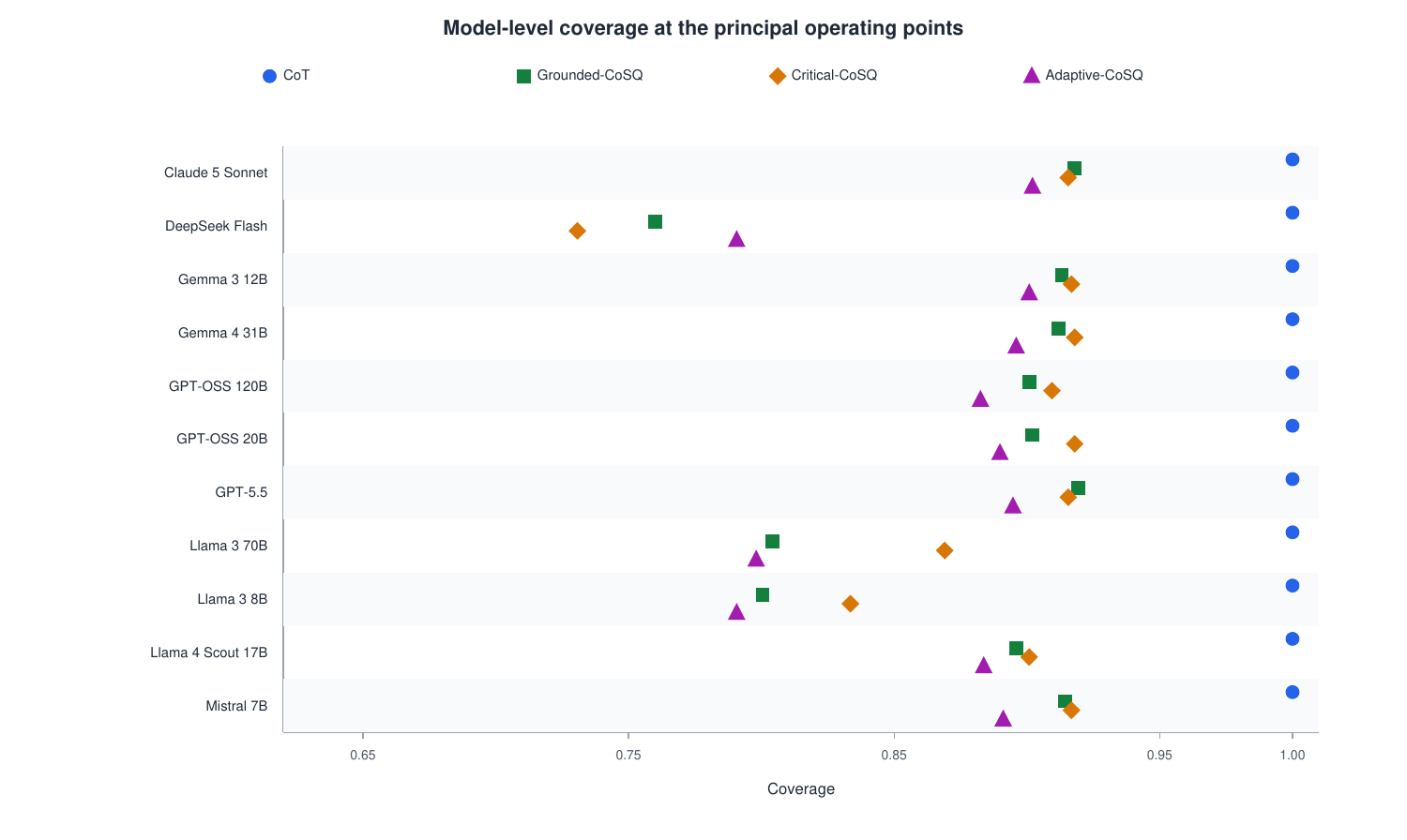}
\caption{Model-level coverage for CoT and the three $\tau=0.90$ CoSQ variants. Coverage is an operating characteristic: lower values indicate that more questions are explicitly referred rather than answered.}
\label{fig:modelcoverage}
\end{figure*}

The aggregate result is not driven by a single model. At $\tau=0.90$, all three CoSQ variants have lower HR and higher AA than CoT for every model in the panel. Grounded-CoSQ HR reductions range from 1.84 to 9.79 percentage points; the corresponding ranges are 1.71--10.28 points for Critical-CoSQ and 2.45--8.94 points for Adaptive-CoSQ. DeepSeek Flash shows the largest absolute reduction under Grounded-CoSQ, from 0.177 to 0.080, while Critical-CoSQ provides the lowest HR for both DeepSeek Flash and Llama 3 70B. Table~\ref{tab:modelresults} reports the complete model-level values rather than only a pooled mean. Figures~\ref{fig:modelhr} and~\ref{fig:modelcoverage} separate the corresponding HR and coverage profiles to preserve readability.

\begin{table*}[t]
\centering
\caption{Model-level TruthfulQA-MC results for Direct, CoT, and the three principal selective operating points. Each cell reports AA / coverage / HR, where HR is the unconditional wrong-commitment rate $W/N$.}
\label{tab:modelresults}
\scriptsize
\resizebox{\linewidth}{!}{%
\begin{tabular}{lccccc}
\toprule
Model & Direct & CoT & Grounded-CoSQ $\tau=0.90$ &
Critical-CoSQ $\tau=0.90$ & Adaptive-CoSQ $\tau=0.90$ \\
\midrule
Llama 3 8B
& 0.818/1.000/0.182
& 0.807/1.000/0.193
& 0.849/0.800/0.121
& 0.838/0.834/0.135
& 0.853/0.791/0.116 \\

Llama 3 70B
& 0.814/1.000/0.186
& 0.803/1.000/0.197
& 0.865/0.804/0.109
& 0.882/0.869/0.103
& 0.842/0.798/0.126 \\

Llama 4 Scout 17B
& 0.892/1.000/0.108
& 0.887/1.000/0.113
& 0.910/0.896/0.081
& 0.901/0.901/0.089
& 0.902/0.884/0.087 \\

Gemma 3 12B
& 0.892/1.000/0.108
& 0.892/1.000/0.108
& 0.909/0.913/0.083
& 0.904/0.917/0.088
& 0.917/0.901/0.075 \\

Gemma 4 31B
& 0.890/1.000/0.110
& 0.892/1.000/0.108
& 0.907/0.912/0.084
& 0.908/0.918/0.084
& 0.914/0.896/0.077 \\

Mistral 7B
& 0.891/1.000/0.109
& 0.891/1.000/0.109
& 0.906/0.914/0.086
& 0.900/0.917/0.092
& 0.916/0.891/0.075 \\

GPT-OSS 20B
& 0.890/1.000/0.110
& 0.891/1.000/0.109
& 0.906/0.902/0.084
& 0.907/0.918/0.086
& 0.909/0.890/0.081 \\

GPT-OSS 120B
& 0.892/1.000/0.108
& 0.882/1.000/0.118
& 0.913/0.901/0.078
& 0.896/0.909/0.094
& 0.895/0.882/0.093 \\

GPT-5.5
& 0.890/1.000/0.110
& 0.895/1.000/0.105
& 0.905/0.919/0.087
& 0.910/0.916/0.082
& 0.910/0.895/0.081 \\

Claude 5 Sonnet
& 0.892/1.000/0.108
& 0.892/1.000/0.108
& 0.905/0.918/0.087
& 0.912/0.916/0.081
& 0.912/0.902/0.080 \\

DeepSeek Flash
& 0.835/1.000/0.165
& 0.823/1.000/0.177
& 0.895/0.760/0.080
& 0.898/0.731/0.075
& 0.889/0.791/0.088 \\
\bottomrule
\end{tabular}%
}
\end{table*}

\subsection{Paired statistical analysis}

Table~\ref{tab:stats} reports the paired model-level analysis. For the prespecified Grounded-CoSQ comparison, mean HR decreases by 0.0422 (95\% bootstrap interval [0.0272, 0.0595]) and AA increases by 0.0287 ([0.0177, 0.0418]). Both directional Wilcoxon tests yield $p=0.00049$, and all eleven model-level differences have the hypothesized sign. H1 and H2 are therefore supported on the final balanced TruthfulQA-MC panel. Critical-CoSQ and Adaptive-CoSQ show the same directional consistency, with HR reductions of 0.0396 and 0.0424 and AA gains of 0.0273 and 0.0275, respectively.

\begin{table*}[t]
\centering
\caption{Paired model-level changes relative to CoT at $\tau=0.90$. Positive values denote improvement: lower HR or higher AA. The analysis uses eleven models, 10,000 bootstrap resamples, exact one-sided Wilcoxon signed-rank tests, and paired-sample Cohen's $d_z$.}
\label{tab:stats}
\scriptsize
\begin{tabular}{llrcrrr}
\toprule
Comparison & Outcome & Mean change & Bootstrap 95\% CI & Wilcoxon $p$ & $d_z$ & Direction \\
\midrule
Grounded-CoSQ & HR reduction & 0.0422 & [0.0272, 0.0595] & 0.00049 & 1.44 & 11/11 \\
Grounded-CoSQ & AA increase & 0.0287 & [0.0177, 0.0418] & 0.00049 & 1.35 & 11/11 \\
Critical-CoSQ & HR reduction & 0.0396 & [0.0233, 0.0586] & 0.00049 & 1.27 & 11/11 \\
Critical-CoSQ & AA increase & 0.0273 & [0.0149, 0.0431] & 0.00049 & 1.08 & 11/11 \\
Adaptive-CoSQ & HR reduction & 0.0424 & [0.0300, 0.0571] & 0.00049 & 1.75 & 11/11 \\
Adaptive-CoSQ & AA increase & 0.0275 & [0.0191, 0.0376] & 0.00049 & 1.66 & 11/11 \\
\bottomrule
\end{tabular}
\end{table*}

\begin{figure*}[t]
\centering
\includegraphics[width=0.98\linewidth]{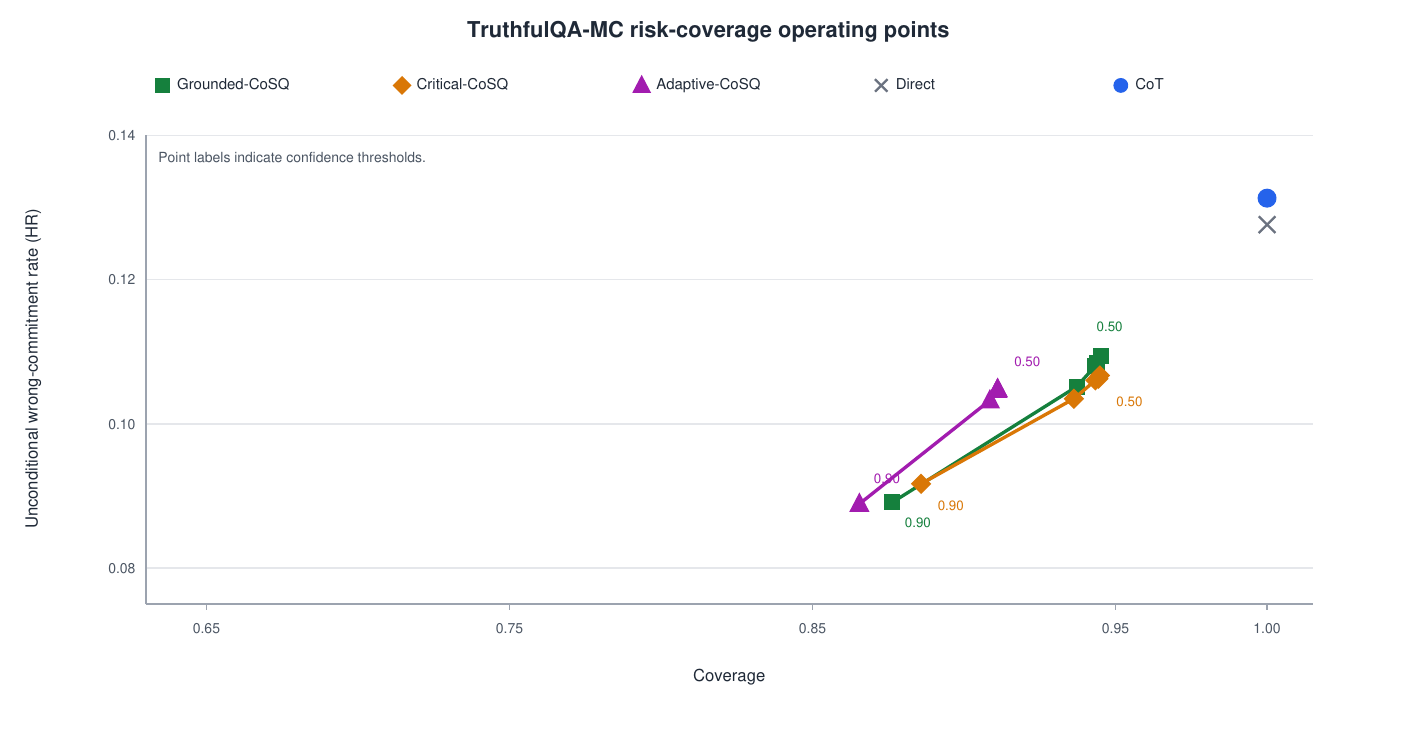}
\caption{TruthfulQA-MC wrong-commitment--coverage operating points. Connected markers show each CoSQ family from $\tau=0.50$ to $0.90$; isolated markers show the forced Direct and CoT baselines. Coverage is reported as a tunable operating characteristic rather than an error measure.}
\label{fig:riskcoverage}
\end{figure*}

The small $p$-values indicate highly consistent paired directions within this panel; they do not make the hosted model endpoints independent samples from a population of all LLMs. The full threshold and model tables therefore remain the primary evidence. Figure~\ref{fig:riskcoverage} summarizes the resulting wrong-commitment--coverage operating points.

\clearpage
\section{Discussion}

The results support a precise interpretation of CoSQ as a selective decision mechanism. At $\tau=0.90$, all three variants increased AA and reduced unconditional wrong commitment relative to CoT for every model in the panel. Grounded-CoSQ reduced mean HR from 0.131 to 0.089, a 4.22-point absolute and 32.1\% relative reduction, while increasing AA from 0.869 to 0.897. Its coverage of 0.876 is an intended operating characteristic: 12.3\% of questions were directed to abstention instead of receiving a potentially unsupported commitment. The abstention-composition analysis makes this selectivity transparent. Some abstentions prevent CoT errors, whereas others withhold answers that CoT happens to answer correctly; the latter are not scoring errors, but they quantify the reduction in automated service breadth at this operating point. Critical-CoSQ offers the broadest $\tau=0.90$ profile, covering 0.886 of questions while retaining a 3.96-point HR reduction and a 2.73-point AA increase. Adaptive-CoSQ is more conservative, covering 0.865 and attaining the lowest mean HR, 0.0889. Grounded-CoSQ nevertheless answers one percentage point more questions than Adaptive-CoSQ, reaches the highest mean AA, and differs in HR by only 0.0002.

The complete threshold sweep guards against selecting an apparently favorable threshold after observing the test results. Every one of the 15 selective configurations had lower model-macro HR than CoT, although the improvement was modest at lower thresholds and largest at $\tau=0.90$. Thus, the sweep does not imply a single universally optimal threshold. It exposes a family of operating points from which a system designer can choose according to the relative consequences of wrong commitments and abstentions. The conservative point is justified when an unsupported answer is more costly than escalation; applications that value broader automated service can use a lower threshold or Critical-CoSQ. The smaller HR separation in the MC experiment than in the earlier open-form pilot is also plausible under a ceiling effect: deterministic option scoring leaves less room for a prompting intervention to improve correctness. This interpretation is provisional because the two protocols differ in more than difficulty and are not pooled statistically.

The model-level results do not support the claim that a larger or newer model automatically produces substantially higher CoSQ coverage. Instead, model families differ in how their self-assessed scores map to the gate, while the risk reduction is directionally stable. This suggests that selective answering should remain an explicit system component even when the underlying model is strong.

The use of multiple-choice TruthfulQA is a deliberate methodological improvement. Earlier open-form evaluations can confound factual correctness with answer parsing. In the final protocol, option positions are deterministically balanced within each option-count stratum, and Direct and CoT are scored as forced-choice baselines: malformed outputs are wrong rather than abstentions. This removes fixed-label and accidental-abstention explanations for the primary result. The limitation is that multiple-choice evaluation does not represent every property of open-ended factual dialogue. For this reason, NQ-Short is treated as secondary generalization evidence rather than folded into the confirmatory estimate.

Grounded-CoSQ is especially relevant when an answer can initiate a costly action. In healthcare, legal assistance, and other high-stakes domains, the framework should not be interpreted as a substitute for professional review. Its practical role is to reduce unsupported commitments and create a clear escalation path for uncertain cases. The correct deployment objective is domain-specific: a system should choose the threshold using the relative costs of false answers, abstentions, and missed opportunities.

\section{Limitations}

First, confidence is elicited from the same language model that produces the answer. It is therefore not an independent probability estimate and may be miscalibrated. The threshold values are empirical operating points, not literal probabilities.

Second, the model panel uses endpoint-level identifiers and hosted model names. Some endpoints may change weights or serving behavior without exposing immutable revisions. The run manifests, prompt versions, and aggregate tables are therefore essential to interpreting the results.

Third, CoSQ adds inference calls and token cost because it decomposes a question and evaluates information units before answering. This cost is justified only when the expected cost of a wrong answer is sufficiently high. The reported benefits should therefore be interpreted together with Table~\ref{tab:cost}, and future work should compare methods under an explicit cost budget.

Fourth, TruthfulQA-MC is the primary benchmark and NQ-Short is a smaller secondary evaluation. Although the two datasets provide complementary evidence, larger multi-domain evaluations are required before claiming broad deployment validity. Threshold transfer across datasets and domains is also unresolved.

Fifth, the analysis uses a common prompt protocol rather than model-specific prompt optimization. This improves comparability but may leave performance on the table for some models. The study also does not exhaustively test paraphrases, instruction order, or alternative confidence wording. Prompt robustness is therefore an open limitation rather than an assumption of invariance. Human assessment of abstention quality and answer usefulness is another important next step.

\section{Conclusion}

This paper introduced Chain-of-Self-Questioning as a prompt-only framework for selective factual answering. Across eleven models and 817 TruthfulQA-MC questions with balanced option positions and forced-choice baselines, Grounded-CoSQ at $\tau=0.90$ reduced the unconditional wrong-commitment rate relative to CoT by 4.22 percentage points and increased answered accuracy by 2.87 points while retaining 87.6\% coverage. Critical-CoSQ retained broader coverage with a 3.96-point HR reduction and a 2.73-point AA increase. Adaptive-CoSQ achieved the lowest mean HR, reducing it by 4.24 points while increasing AA by 2.75 points. All three directional effects held for every model in the panel and were supported by paired exact Wilcoxon tests. The complete threshold sweep showed that these configurations are operating choices on a risk--coverage frontier rather than fixed claims of universal superiority.

The central lesson is that reliable LLM evaluation should ask two questions:
how accurate are the answers that the model gives, and how often does it
decide to give an answer? CoSQ makes that distinction operational.
Grounded-CoSQ at $\tau=0.90$ provides the strongest balance of answered
accuracy, coverage, and wrong-commitment risk in these experiments. Critical-CoSQ is preferable when broader answer coverage is required, whereas Adaptive-CoSQ provides the lowest mean wrong-commitment rate. The framework offers a lightweight
mechanism for reducing unsupported commitments in settings where saying
``I do not know'' is preferable to producing a fluent but false answer.

\bmsubsection*{Data Availability Statement}

The experimental framework, configuration examples, aggregate tables, and
figure-generation scripts are publicly available at
\url{https://github.com/senolali/CoSQ}. The package is also distributed
through PyPI at \url{https://pypi.org/project/cosq}. The exact grounded,
critical-item, and adaptive prompt templates used in the experiments are
reproduced in Appendix~\ref{app:prompts}, and the model-level NQ-Short
results are provided in Appendix~\ref{app:nqmodel}. Versioned source files
and aggregate results are supplied with the reproducibility package.
Provider credentials, private endpoints, local caches, and operational
configuration files are excluded from the public release.

\bmsubsection*{Ethics Statement}

The study uses public benchmark data and does not involve human subjects or private personal data. CoSQ is not a guarantee of factual correctness and should not be deployed in high-stakes settings without domain-specific validation, evidence access, and human oversight. Abstention thresholds should be selected using the cost of wrong answers and unanswered questions in the target application.

\bmsubsection*{Author Contributions}

Ali \c{S}enol conceived the study, developed the method and software, conducted the experiments and statistical analyses, prepared the visualizations, and wrote and revised the manuscript.

\bmsubsection*{Conflicts of Interest}

The author declares no conflict of interest.

\bmsubsection*{Acknowledgments}

This work was supported by the Scientific and Technological Research Council of T\"urkiye (T\"UB\.ITAK) under Project No. 126E534.

\clearpage

\clearpage
\onecolumn
\appendix

\begingroup
\raggedright
\fvset{
  fontsize=\small,
  breaklines=true,
  breakanywhere=true,
  breaksymbolleft={},
  breaksymbolright={}
}

\section{Prompt Templates}
\label{app:prompts}

The templates below reproduce the versioned prompt files used in the
TruthfulQA-MC and NQ-Short experiments. The placeholder
\texttt{\{question\_block\}} contains the rendered question and, for the
multiple-choice protocol, its answer options. The Direct and CoT baselines
in the multiple-choice experiment are forced-choice conditions and do not
offer abstention. The CoT comparator is an instruction-level reasoning
baseline: the model is instructed to reason silently, while only the final
option label is recorded. A separate CoT-with-abstention condition was not
included in the reported 17-condition experiment and is therefore not
reproduced here.

The confidence thresholds, aggregation rules, and acceptance decisions are
implemented by the framework and are not additional prompt text. When a
CoSQ gate rejects a question, the framework records the common abstention
output shown at the end of this appendix.
\Needspace{8\baselineskip}
\subsection{Question wrappers}

\noindent\textbf{Multiple-choice wrapper
(\texttt{question.v1.txt}).}
\begin{Verbatim}
Question: {question}

Options:
{options}
\end{Verbatim}

\medskip
\noindent\textbf{Open-form wrapper
(\texttt{question\_open.v1.txt}).}
\begin{Verbatim}
Question: {question}
\end{Verbatim}

\Needspace{8\baselineskip}
\subsection{Multiple-choice baselines}

\noindent\textbf{Forced-choice Direct
(\texttt{direct\_mc.v1.txt}).}
\begin{Verbatim}
{question_block}
Return exactly one token corresponding to one of the option labels shown above. Do not provide an explanation.
\end{Verbatim}

\medskip
\noindent\textbf{Forced-choice CoT
(\texttt{cot\_mc.v1.txt}).}
\begin{Verbatim}
{question_block}
Think silently and return exactly one token corresponding to one of the option labels shown above. Do not provide reasoning or explanation.
\end{Verbatim}

\Needspace{8\baselineskip}
\subsection{Open-form baselines}

\noindent\textbf{Open-form Direct
(\texttt{direct\_open.v1.txt}).}
\begin{Verbatim}
{question_block}
Answer in one short sentence.

Answer:
\end{Verbatim}

\medskip
\noindent\textbf{Open-form CoT
(\texttt{cot\_open.v1.txt}).}
\begin{Verbatim}
{question_block}
Let's think step by step, then give your final answer on a line beginning with
"Answer:". Keep the final answer to one short sentence.

Reasoning:
\end{Verbatim}

\Needspace{8\baselineskip}
\subsection{Grounded-CoSQ information-unit stage}

\noindent\textbf{Information-unit identification
(\texttt{cosq\_needs.v1.txt}).}
This template is shared by Grounded-CoSQ and Adaptive-CoSQ.
\begin{Verbatim}
{question_block}
Before answering, list the individual pieces of information you would need in order to
answer this question correctly. Do not answer the question yet.

List them one per line, numbered, with no commentary.

Required information:
\end{Verbatim}

\Needspace{8\baselineskip}
\subsection{Grounded-CoSQ claim and confidence stage}

\noindent\textbf{Factual claim and confidence elicitation
(\texttt{cosq\_fact\_confidence.v1.txt}).}
The template is applied separately to every identified information need.
\begin{Verbatim}
{question_block}
Consider this information need:

{need}

State the factual claim you would rely on to answer the question.
Then give your confidence in that claim as an integer from 0 to 100.
Use exactly this format:
FACT: <one concise factual claim>
CONFIDENCE: <0-100>
Do not answer the main question yet.
\end{Verbatim}

\Needspace{8\baselineskip}
\subsection{Grounded-CoSQ answer stage}

\noindent\textbf{Multiple-choice answer
(\texttt{cosq\_grounded\_answer\_mc.v1.txt}).}
\begin{Verbatim}
{question_block}
The following factual claims passed the confidence gate:

{accepted_facts}

Return exactly one token corresponding to one of the option labels shown above, or ABSTAIN. Do not provide reasoning or explanation.
\end{Verbatim}

\medskip
\noindent\textbf{Open-form answer
(\texttt{cosq\_grounded\_answer.v1.txt}).}
\begin{Verbatim}
{question_block}
The following factual claims passed the confidence gate:

{accepted_facts}

Use these claims as evidence, reason briefly, and give the final answer on a line
beginning with "Answer:". Keep the final answer to one short sentence.

Reasoning:
\end{Verbatim}

\Needspace{8\baselineskip}
\subsection{Critical-CoSQ information classification}

\noindent\textbf{Critical and supporting information units
(\texttt{cosq\_needs\_roles.v1.txt}).}
\begin{Verbatim}
{question_block}
Before answering, list the individual pieces of information you would need to answer
this question correctly. Classify every item as either critical or supporting.

CRITICAL means that without this information, no correct answer is possible.
SUPPORTING means that the information is helpful but the question could still be
answered without it.

Use exactly one item per line in this format:
[CRITICAL] <information needed>
or
[SUPPORTING] <information needed>

Do not answer the main question yet.
Required information:
\end{Verbatim}

\Needspace{8\baselineskip}
\subsection{Critical-CoSQ claim and confidence stage}

\noindent\textbf{Critical claim and confidence elicitation
(\texttt{cosq\_critical\_fact\_confidence.v1.txt}).}
The template is applied separately to every information unit classified as
critical.
\begin{Verbatim}
{question_block}
This is a critical information need:

{need}

State the factual claim you would rely on and give your confidence in that claim.
Use exactly this format:
FACT: <one concise factual claim>
CONFIDENCE: <0-100>
Do not answer the main question yet.
\end{Verbatim}

\Needspace{8\baselineskip}
\subsection{Critical-CoSQ answer stage}

\noindent\textbf{Multiple-choice answer
(\texttt{cosq\_critical\_grounded\_answer\_mc.v1.txt}).}
\begin{Verbatim}
{question_block}
The following critical factual claims passed the confidence gate:

{accepted_facts}

Use these claims as evidence. Return exactly one token corresponding to one of the option labels shown above, or ABSTAIN. Do not provide reasoning or explanation.
\end{Verbatim}

\medskip
\noindent\textbf{Open-form answer
(\texttt{cosq\_critical\_grounded\_answer.v1.txt}).}
\begin{Verbatim}
{question_block}
The following critical factual claims passed the confidence gate:

{accepted_facts}

Use these claims as evidence and do not contradict them. You may use general reasoning.
Give the final answer on a line beginning with "Answer:" and keep it to one short sentence.

Reasoning:
\end{Verbatim}

\Needspace{8\baselineskip}
\subsection{Adaptive-CoSQ role-aware assessment}

\noindent\textbf{Role, factual claim, and confidence elicitation
(\texttt{cosq\_fact\_role\_confidence.v1.txt}).}
The template is applied separately to every information unit identified by
the shared information-unit stage.
\begin{Verbatim}
{question_block}
Consider this information need:

{need}

State the factual claim you would rely on. Mark whether it is critical or supporting,
then give confidence from 0 to 100.
Use exactly this format:
ROLE: critical or supporting
FACT: <one concise factual claim>
CONFIDENCE: <0-100>
Do not answer the main question yet.
\end{Verbatim}

\Needspace{8\baselineskip}
\subsection{Adaptive-CoSQ answer stage}

\noindent\textbf{Multiple-choice answer
(\texttt{cosq\_grounded\_adaptive\_answer\_mc.v1.txt}).}
\begin{Verbatim}
{question_block}
The following factual claims passed the adaptive confidence checks:

{accepted_facts}

Return exactly one token corresponding to one of the option labels shown above, or ABSTAIN. Do not provide reasoning or explanation.
\end{Verbatim}

\medskip
\noindent\textbf{Open-form answer
(\texttt{cosq\_grounded\_adaptive\_answer.v1.txt}).}
\begin{Verbatim}
{question_block}
Accepted factual claims:

{accepted_facts}

Answer mode: {response_mode}
Use the claims as evidence and do not contradict them. You may use general reasoning.
Give the final answer on a line beginning with "Answer:" and keep it to one short sentence.
For cautious mode, briefly signal uncertainty without refusing if the evidence supports
a useful answer.

Reasoning:
\end{Verbatim}

\Needspace{8\baselineskip}
\subsection{Adaptive-CoSQ consistency check}

\noindent\textbf{Post-answer contradiction check
(\texttt{cosq\_consistency\_check.v1.txt}).}
\begin{Verbatim}
{question_block}
Accepted factual claims:
{accepted_facts}

Candidate answer:
{answer}

Does the candidate answer contradict any accepted factual claim?
Reply with exactly one word: CONSISTENT or CONTRADICTORY.
\end{Verbatim}

\Needspace{8\baselineskip}
\subsection{Shared abstention output}

When a CoSQ gate rejects a question, or when the Adaptive-CoSQ consistency
check identifies a contradiction, the framework records the following
output without making another answer-generation call
(\texttt{cosq\_abstain.v1.txt}).

\begin{Verbatim}
I don't know.
\end{Verbatim}
\endgroup

\clearpage
\section{Supplementary NQ model-level results}
\label{app:nqmodel}
Table~\ref{tab:nqmodel} provides the complete model-level values underlying the aggregate NQ-Short results in Table~\ref{tab:nq}. Each cell reports AA / coverage / HR. NQ was run with five representative models and the current $\tau=0.90$ operating point; no NQ threshold sweep is used for the primary claims.

\begin{table}[htbp]
\centering
\caption{Model-level NQ-Short results. Each cell reports answered accuracy / coverage / unconditional wrong-commitment rate.}
\label{tab:nqmodel}
\scriptsize
\resizebox{\linewidth}{!}{%
\begin{tabular}{lccccc}
\toprule
Model & Direct & CoT & Grounded-CoSQ $\tau=0.90$ & Critical-CoSQ $\tau=0.90$ & Adaptive-CoSQ $\tau=0.90$ \\
\midrule
Llama 3 8B & 0.613/1.000/0.387 & 0.600/1.000/0.400 & 0.665/0.817/0.273 & 0.632/0.907/0.333 & 0.664/0.863/0.290 \\
Llama 4 Scout 17B & 0.590/1.000/0.410 & 0.583/1.000/0.417 & 0.688/0.833/0.260 & 0.638/0.893/0.323 & 0.663/0.860/0.290 \\
Gemma 4 31B & 0.607/1.000/0.393 & 0.590/1.000/0.410 & 0.673/0.837/0.273 & 0.626/0.883/0.330 & 0.636/0.870/0.317 \\
GPT-OSS 20B & 0.587/1.000/0.413 & 0.590/1.000/0.410 & 0.672/0.813/0.267 & 0.629/0.917/0.340 & 0.654/0.867/0.300 \\
GPT-5.5 & 0.590/1.000/0.410 & 0.577/1.000/0.423 & 0.651/0.850/0.297 & 0.662/0.877/0.297 & 0.653/0.873/0.303 \\
\bottomrule
\end{tabular}}
\end{table}

\clearpage

\end{document}